\documentclass[11pt]{article}

\usepackage[preprint]{acl}

\usepackage{times}
\usepackage{latexsym}
\usepackage{subcaption}
\usepackage{xcolor}
\definecolor{cbBlue}{RGB}{0,114,178}
\definecolor{cbOrange}{RGB}{230,159,0}
\definecolor{cbGreen}{RGB}{0,158,115}
\definecolor{cbPurple}{RGB}{204,121,167}
\definecolor{cbYellow}{RGB}{255,215,0}
\usepackage{comment}

\usepackage[T1]{fontenc}

\usepackage[utf8]{inputenc}

\usepackage{microtype}

\usepackage{inconsolata}

\usepackage{graphicx}
\usepackage{booktabs}
\usepackage{placeins}   
\usepackage{dblfloatfix} 

\title{Conspiracy Frame: a Semiotically-Driven Approach for Conspiracy Theories Detection}

\author{
 \textbf{Heidi Campana Piva\textsuperscript{1}},
 \textbf{Shaina Ashraf\textsuperscript{2}},
 \textbf{Maziar Kianimoghadam Jouneghani\textsuperscript{1}},
 \\
  \textbf{Arianna Longo\textsuperscript{1}},
  \textbf{Rossana Damiano\textsuperscript{1}},
  \textbf{Lucie Flek\textsuperscript{2}},
  \textbf{Marco Antonio Stranisci\textsuperscript{1,3}},
\\
\\
\\
 \textsuperscript{1}University of Turin, Italy
 \textsuperscript{2}University of Bonn, Germany
  \textsuperscript{3}aequa-tech, Italy
   \\
  \small{
   \textbf{Correspondence:} \href{mailto:marcoantonio.stranisci@unito.it}{marcoantonio.stranisci@unito.it} \href{mailto:heidi.campanapiva@unito.it}{heidi.campanapiva@unito.it}
  }
}

\begin{document}
\maketitle
\begin{abstract}
Conspiracy theories are anti-authoritarian narratives that lead to social conflict, impacting how people perceive political information. To help in understanding this issue, we introduce the Conspiracy Frame: a fine-grained semantic representation of conspiratorial narratives derived from frame-semantics and semiotics, which spawned the \textbf{Con}spiracy \textbf{Fra}mes (Con.Fra.) dataset: a corpus of Telegram messages annotated at span-level. The Conspiracy Frame and Con.Fra. dataset contribute to the implementation of a more generalizable understanding and recognition of conspiracy theories. We observe the ability of LLMs to recognize this phenomenon in-domain and out-of-domain, investigating the role that frames may have in supporting this task. Results show that, while the injection of frames in an in-context approach does not lead to clear increase of performance, it has potential; the mapping of annotated spans with FrameNet shows abstract semantic patterns (e.g., `Kinship', `Ingest\_substance') that potentially pave the way for a more semantically- and semiotically-aware detection of conspiratorial narratives.
\end{abstract}

\section{Introduction}

Conspiracy theories (CTs) are narratives that embody an anti-authoritarian stance and skepticism towards official discourses, leading to social discord and a fractured sense of truth \cite{[3],[4],[5],[6]}. CTs play a significant role in the propagation of Information Disorder \citep{wardle2017information} as they shape people's interpretation of reality \cite{[3]} and drive behaviors that have serious economic and political repercussions. Implementing effective strategies to mitigate CTs' negative impact on society is becoming a priority of many societal actors who give impulse to educational \citep{jerome2024combatting} and research \citep{costello2024durably} initiatives to combat them.

Research in Natural Language Processing (NLP) on CTs is crucial, since it enables automatic recognition and potential countering of this phenomenon through the development of datasets \citep{pogorelov2021wico,phillips2022hoaxes} and novel approaches to detect and characterize CTs \citep{corso2025conspiracy,schlicht2025detection}. However, two main challenges need to be addressed to implement effective technologies against this phenomenon: 1\textsuperscript{st}- From a computational perspective, CTs are hard to define and, thus, to annotate. Almost all existing corpora are annotated for binary and multi-label classification, which may not be useful to distinguish CTs from other information disorders like disinformation \citep{da2020survey} and propaganda \citep{hussain2025fake}, since they do not provide information on the structure of these narratives. 2\textsuperscript{nd}- From a thematic perspective, NLP resources for this task suffer a lack of generalizability, because most of them are confined to COVID-19, vaccines, or other health-related topics and restricted to particular contents.

The aim of this work is to \textbf{support a more general understanding of CTs to improve automatic recognition of this phenomenon}. To this extent, we introduce the Conspiracy Frame: a fine-grained semantic representation of conspiratorial narratives whose design derives from frame-semantics and semiotics \citep{fillmore1976frame}. The Conspiracy Frame translates the definition of CTs from semiotics \citep{[3],[10]} into a narrative structure that involves the presence of five frame elements: \textit{plan}, \textit{secret}, \textit{call-to-action}, \textit{in-group}, and \textit{out-group}. This semantic representation drove the creation of the \textbf{Con}spiracy \textbf{Fra}mes (Con.Fra.) dataset, a corpus of Telegram messages annotated at span-level according to the Conspiracy Frame. The dataset is not only conceived for the detection of CTs, but to foster a more general understanding of CTs by seeking generalizability through the consideration of structural (formal) characteristics that are intrinsic to conspiratorial narratives. 

Recognizing the need to assess the generalizability of CTs, Con.Fra has been created through a three-step iterative annotation process: \textit{i. annotation}: we annotated $2,077$ Telegram messages for CTs identification and span detection; \textit{ii. augmentation}: we performed a lexical mapping of annotated spans with frames included in FrameNet \citep{baker1998berkeley} to observe the presence of more abstract semantic patterns emerging from conspiratorial narratives; \textit{iii. out-of-domain evaluation}: we evaluated the ability of LLMs to identify CTs and detect CT spans on messages from different Telegram channels.

The Conspiracy Frame and Con.Fra. dataset contribute to the implementation of more explainable and generalizable CT understanding and recognition. We exploit this novel set of resources to observe the ability of LLMs to recognize this phenomenon in-domain and out-of-domain and to study the role that frames may have in supporting this task. Results show that, while the injection of frames in an in-context approach does not lead to clear increase of performance, it has a promising role in the development of more explainable approaches to CT recognition. The mapping of annotated spans with FrameNet shows abstract semantic patterns (e.g., `Kinship', `Ingest\_substance') that potentially pave the way for a more semantically- and semiotically-aware detection of conspiratorial narratives.

The paper is structured as follows: Section \ref{sec:related} presents existing work on CTs in NLP and Semiotics; Section \ref{sec:corpus} describes the design of the Conspiracy Frame and the creation of the Con.Fra corpus, followed by Section \ref{sec:experiment} which details our experimental setting. Finally, Section \ref{sec:results} is devoted to the analysis of LLM performances on CT recognition and span detection. 
\section{Related Work} \label{sec:related}

CTs have influenced public discourse since at least the 19th century \cite{[1]}, but their reach and impact have been magnified by digital technology \cite{[2]}. Content-wise, conspiracy themes range from government surveillance, to corporate manipulation of scientific knowledge, to media propaganda, and to supernatural knowledge being suppressed by epistemic gatekeepers \cite{[7]}. The operation of CTs is thus more dependent on a specific rhetoric that symbolically reproduces segregation than on their particular contents \cite{[8]}. Because of that, CTs operate as an ideological lens \cite{[9]}, meaning-making template \cite{[10]}, mode of interpretation \cite{[3]} or belief system \cite{[11]}, shaping reality comprehension.


The CT phenomenon has triggered the emergence of large datasets. For instance, the basis LOCO corpus \cite{miani2021loco} provides 88 million words of conspiracy and mainstream documents, allowing lexical feature analysis. Moreover, WICO Text \cite{pogorelov2021wico} and the Hoaxes and Hidden Agendas dataset \cite{phillips2022hoaxes} target specific topics, such as COVID-19, 5G, and climate change CTs, offering text annotated for stance and topic analyses. More recently, the XAI-DisInfodemics dataset \cite{korencic2024distinguishes} introduced a distinction between 'critical' and 'conspiratorial' narratives on Telegram, and the work by Hoseini et al. \cite{hoseini2023globalization} investigated the globalization of the QAnon conspiracy theory on the same platform.

Despite the importance of these resources and tools for training binary classifiers (e.g., \cite{schlicht2025detection, corso2025tiktok}), they typically treat conspiracy theories as a classification label or a topic. However, recent studies on annotation reliability suggest that this binary approach is fraught with subjectivity. \citet{golbeck2018fake} argued the nuanced linguistic differences required to distinguish fake news from satire. \citet{mompelat2022loco} found that distinguishing conspiracy texts from mainstream ones is difficult without precise definitions of belief manifestation. Similarly, \citet{hemm2024serious} demonstrated that even experts struggle to agree on binary labels, showing that the complexity of conspiratorial discourse requires a more nuanced, structural representation than simple labels can provide.
For this reason, the use of semiotics as a methodological perspective for the study of CT is convenient. Past works have considered the semio-technological structures and codes of CTs \cite{[15]}; their logic of signification \cite{[16]}; CT as communicative phenomenon (re)constructing social realities \cite{[4]}, and as strategic narrative \cite{[17]}.

The current work belongs to the intersection of computational CT detection and frame semantic theory. Despite the effectiveness in CT identification, the current state-of-the-art solutions tend to disregard the semiotic structure in CTs. At the same time, the frame semantic approach proved to be quite effective in the fields of disinformation and toxic language but has not been adapted to the specific CT structure yet (considering form, not content). 

\section{The Conspiracy Frames Corpus} \label{sec:corpus}

In this section, we describe the process of creating the corpus of Telegram posts annotated for the fine-grained detection and understanding of CT structure (Con.Fra.).

\subsection{Data collection}
Given the demonstrated role of Telegram  in the propagation of disinformation and conspiracy theories \citep{hoseini2023globalization}, we selected it as a target for our data collection\footnote{In order to keep the researchers involved in this work as well as the Telegram users safe from any potential harassment, the names of the channels have been anonymized}, which involved two steps: \textit{i.} we performed a qualitative selection of public Telegram groups to design the annotation scheme; \textit{ii.} we then operated a massive crawling of all the conspiratorial Telegram groups, starting from the list provided by \citet{[12]}. As a result, we obtained a total of $11,618$ channels.

The channels selected from this list were chosen to be as diversified as possible, ensuring that the data were not biased toward a particular demographic group. By ordering the winnowed channels according to the number of subscribers, we selected the Top-35 and, in order to observe geopolitical variation, we gathered all the messages published in these groups during a period of two years before the start of the research (2023 and 2024). 

\subsection{The Annotation Scheme: Designing the Conspiracy Frame}
The direct adoption of FrameNet \citep{baker1998berkeley} for the annotation of conspiratorial narratives was not feasible for the annotation scheme since there are no existing frames specifically designed for 'conspiracy theory' and because the token-level annotation procedure requires expert annotators in the field of Information Extraction. We chose this framework as a source of inspiration to design the \textbf{Conspiracy Frame}, which includes a set of five elements to annotate at span-level.

We derived the conspiracy frame from the definition of CT given by semiotics literature \citep{[3],[10]} : 

\begin{quote}
\colorbox{cbOrange}{a narrative} that represents a determined circumstance as being \colorbox{cbBlue}{the result of a secret plan} implemented by a \colorbox{cbGreen}{morally evil group of people} to hurt a group of \colorbox{cbGreen}{victims}, and that, \colorbox{cbPurple}{if left unstopped}, will lead to catastrophe. 
\end{quote}

To design frame elements, we used semiotics as a foundation, considering four interrelated levels of CT analysis \cite{[10]}: \textit{i.} CT analysed as representation (\colorbox{cbOrange}{a narrative}); \textit{ii.} CT analysed as mode of interpretation (\colorbox{cbBlue}{the result of a secret plan.}); \textit{iii.} analysis of the processes of identity construction in CT (\colorbox{cbGreen}{evil people}/\colorbox{cbGreen}{victims}); and \textit{iv.} analysis of the call-to-action (\colorbox{cbPurple}{what must be done to hinder catastrophe})

We take conspiracy theory as a discursive event (as an instance of language use) and, more specifically, as \textit{political discourse} - which is premised on the capacity of actors to drive change \cite{[13]}. From this perspective, discourse as representation \textit{i.} comprises premises of practical arguments (`secret plan') and serve as reasons for action (`call-to-action'), not only describing what social reality is \textit{ii.}, but also what it should be.

Lastly, we highlight that the components of a discursive event cannot exist apart from it. In semiotics, a sign is not intelligible in isolation - it is relational in nature, needing the existence of that which it signifies in order for it to make any sense. Along the same lines, although the elements of political discourse can be identified separately, they cannot exist `as being separate' \cite{[14]}; that is, an in-group only exists in relation to an out-group (`us-vs.-them'). With that, the Conspiracy Frame is characterized by the following elements:

Core elements
\begin{itemize}
\itemsep=-0.5em
\item plan: the event(s), circumstance(s), or action(s) being taken to achieve the evil end (e.g., `ethnic replacement', `forced vaccination');
\item secret: gives the connotation of secrecy to the Plan: (e.g., `they are silencing those who say the truth', `the Big Media is misleading');
\end{itemize}
Non-core elements
\begin{itemize}
\itemsep=-0.5em
\item in-group: the victims - ``us'' - those that are harmed by the Plan (e.g., `christians', `native europeans');
\item out-group: the enemy - ``them'' (e.g., `elites', `the media');
\item call-to-action: what should be done to avoid the catastrophic effects of the Plan. (e.g., `we must organize a protest', `we can't let this happen');
\end{itemize}

\subsection{The Annotation Workshop}
In order to create the corpus we organized an Annotation Workshop open to PhD and MA students from any field of research. Adopting a participatory approach to dataset creation \citep{caselli2021guiding,delgado2023participatory}, the workshop foresaw a theoretical introduction to the topic, an annotation session, and a round of discussion on the annotation scheme. The workshop was organized in two sessions in line with the two iterations of our annotation task: \textit{i.} Corpus creation; \textit{ii.} Evaluation of LLMs on out-of-domain predictions.
\subsubsection{First session: corpus creation.} \label{sss:first}
The $13$ participants were asked to annotate three batches of $100$ Telegram messages. Each batch included $20$ messages from five different groups and was annotated by two different pairs of annotators. The final corpus includes $2,077$ annotated text-messages.

In line with existing datasets on highly subjective topics like Hate Speech \citep{sanguinetti2018italian} and sexism \citep{kirk2023semeval}, the corpus shows an imbalance distribution of labels: $20.63\%$ of messages were labeled as conspiratorial and $77.00\%$ as non-conspiratorial. 
We calculated pairwise Cohen's Kappa scores for the conspiracy classification task across all annotator pairs. The analysis reveals moderate inter-annotator agreement with a mean Kappa of $0.41$ (SD=$0.20$). 

For posts identified as conspiratorial, annotators marked spans corresponding to the five frame elements. We evaluated the reliability of these span annotations using Average Cohen’s Kappa ($\kappa$). The results indicate substantial agreement across all frame elements: Plan/Event ($\kappa = 0.808$); Call-to-Action ($\kappa = 0.750$); Out-group ($\kappa = 0.717$); Secret ($\kappa = 0.683$); In-group ($\kappa = 0.633$)

\subsubsection{Second session: LLM evaluation.} \label{sss:second}
This section assesses the ability of LLMs to generalize their knowledge about CTs in the classification and span-annotation of messages gathered from unseen Telegram channels. $10$ students participated to this session, which led to an additional dataset of $500$ messages with $10$ annotations each, as described below.

We automatically annotated $2500$ messages and sampled $250$ items to evaluate LLMs performance in the automatic classification of out-of-domain messages. We asked annotators to label each message as containing or not a conspiracy theory. Annotators' agreement in the evaluation of LLMs' classification, which we calculated through Pairwise Cohen's Kappa score, was 0.23 (SD=0.13), significantly lower than the one obtained during the corpus annotation. This is probably due to the characteristics of the task, which focuses on out-of-domain messages and is centered on the evaluation of models rather than the individuation of a specific phenomenon. $250$ additional messages were used to evaluate the quality of span detection, with a specific focus on the core elements of the frame: \textit{plan} and \textit{secret}. We adopted a best-worst scaling approach \citep{poletto2019annotating} in which we provided the annotators with the spans annotated by $4$ different LLMs and asked them to identify the best and the worst. Annotators' agreement in the evaluation of LLMs' span identification was in line with the previous session: $0.69$ (SD=$0.07$) in the evaluation of the best span; $0.62$ (SD=$0.07$) in the evaluation of the worst span. 

The resulting corpus is composed of $2,577$ messages: $2,077$ annotated for CT recognition and span detection and $500$ for the out-of-domain evaluation of LLMs performance.\footnote{The dataset is available at the following \url{https://anonymous.4open.science/r/conspiracy-frame-3D65/README.md}}

\section{Experimental Setting} \label{sec:experiment}
To study the generalizability of CT recognition and the role of frames in the understanding of this phenomenon, we designed an experimental setting in which spans annotated according to the Conspiracy Frame are aligned with FrameNet \citep{baker1998berkeley}. The alignment supports a set of classification and span detection experiments and a qualitative analysis of frames emerging from human annotations and spans detected by LLMs.

\paragraph{Aligning the Conspiracy Frame with FrameNet.} \label{sss:alignment}
In order to perform the alignment, we selected all the annotation spans in the Con.Fra corpus that trigger one or both core elements of the Conspiracy Frame, \textit{plan} and \textit{secret}. To do so, we adopted a lexical-based approach: we lemmatized the tokens occurring in all the spans triggering a plan and/or a secret with SpaCy\footnote{\url{https://spacy.io/}. We used the \textit{en-core-web-lg} model}, and mapped all lemmas with lexical units from FrameNet v1.7.\footnote{\url{https://framenet.icsi.berkeley.edu/}} Whenever a lexical unit matched a lemma in a span we aligned the span with its corresponding frame. For instance, considering a span annotated as a Plan containing the phrase `planning a white genocide', the mapping pipeline proceeds as follows: the lemma 'genocide' is mapped with FrameNet Lexical Unit \textit{genocide.n}, which is associated with the frame \textit{Killing}.

Given the fuzziness of FrameNet taxonomy \citep{lonneker2009framenet} and the presence of very general frames triggered by highly recurrent lexical units, we adopted three heuristics to reduce the noise in the mapping: \textit{i.} We only mapped verbs, nouns and adjectives; \textit{ii.} We removed copular, modal, and light verbs~\citep{bonial2020choosing}\footnote{be, try, have, do, make, get, must, should, can, may, might, want} from the mapping; \textit{iii.} We adopted the method proposed by \citet{clauset2009power} to identify and filter out the tail frames in the distribution. This process enabled the mapping of the annotated spans with $54$ unique frames occurring $2526$ times. 

\paragraph{Frame-driven CT recognition.} \label{sss:ct_recognition}
Once the spans were mapped to FrameNet, we designed a CT classification and span detection experiment based on \textit{in-context} learning. The experiment was organized in two different stages. 

We evaluate LLMs on the annotated dataset using LLaMA-3.3 models, which have demonstrated strong zero-shot instruction following and semantic reasoning capabilities for text classification without fine-tuning \citep{touvron2023llamaopenefficientfoundation, dubey2024llama3}. We test two model sizes (8B and 70B) to examine the trade-off between scalability and label consistency, under three prompting setups that incrementally add knowledge: zero-shot, few-shot, and frame-guided. All prompts follow the theoretical framework outlined in Appendix \textit{n}, grounded in semiotics literature.

We selected the zero-shot and frame-guided approaches and annotated $2,500$ messages from $15$ Telegram groups that had not been seen during the creation of the corpus. We then sampled $500$ messages for a second run of human-evaluation to assess the generalizability of models performance to new data.  

\paragraph{Evaluation Metrics.} \label{sss:metrics}
For the evaluation of LLM performances on the annotated corpus, we adopted traditional NLP metrics: \textit{precision}, \textit{recall}, and \textit{F-score}. Specifically, we focused on the tradeoff between precision and recall in the classification of CTs and span detection. For the second evaluation step, we adopted the ELO rating system \citep{glickman1999rating} to compute the probability of a frame-guided approach being preferred by human-annotators over a few-shot approach in the prediction of a CT. Born as a method to rank players in games, ELO is now extensively used to evaluate LLM performances \citep{boubdir2024elo} through pairwise comparison. LLMs are initialized as a player with a ranking of $1,000$ and each of their predictions is treated as a game: depending on the votes provided by human annotators (Cfr \ref{sss:second}) the game either ends in a draw or one of the two models wins and the ELO ranking is updated accordingly. Since ELO depends on the order of the games, for each model pair, we repeated the experiment $1,000$ times with a different randomization of the same set of predictions and counted the number of times one model obtained a higher ranking than another. 

\paragraph{Qualitative analysis of most recurring frames.} We performed a qualitative analysis of the most recurrent frames (and their corresponding lexical units) emerged from the alignment of \textit{plan} and \textit{secret} spans with FrameNet. The analysis, which is built upon semiotic research on CTs, focuses on general semantic patterns emerging from human annotations and the most relevant differences with LLM predictions in an out-of-domain setting.

\section{Results} \label{sec:results}
This section is organized in two parts: Section \ref{ss:detection} describes the results of CT recognition and span detection in- and out-of-domain; Section \ref{ss:qualitative} is devoted to a qualitative analysis of the alignment between the annotated spans and FrameNet, with a focus on the key differences between humans and LLMs.

\subsection{Automatic CT Detection} \label{ss:detection}
Results presented in this section refer to the set of in-context learning experiments described in Section \ref{sec:experiment}.

\begin{figure*}[t]
    \centering
    \includegraphics[width=0.9\textwidth]{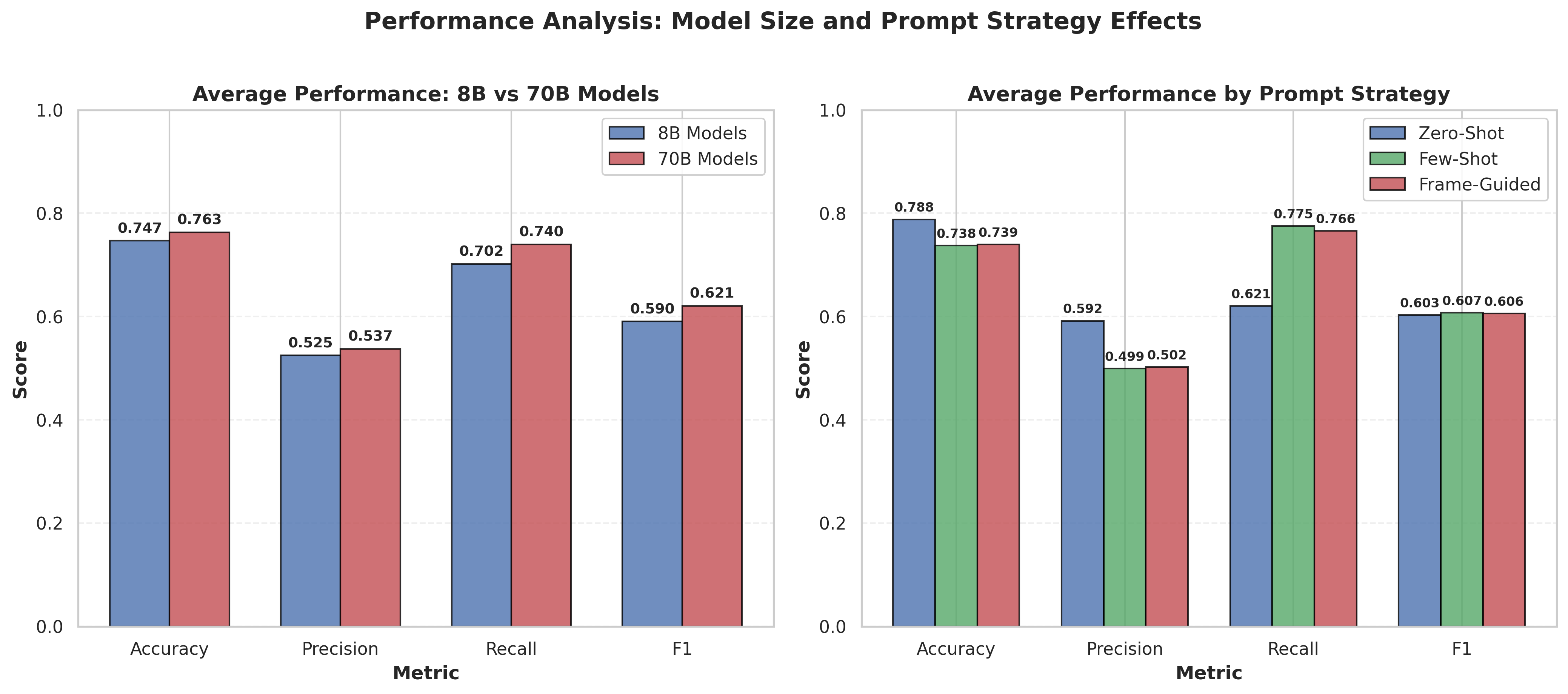}
    \caption{Performance Analysis: Model Size and Prompt Strategy Effects}
    \label{fig:llm_size_strategy}
\end{figure*}
\paragraph{Models' performance on in-domain data.} A first finding of our experiment is that model size has only a moderate impact on CT classification performance: the 70B Llama 3.3 models outperform their 8B counterparts by $3.1$ F-score and $3.8$ recall points  (Figure~\ref{fig:llm_size_strategy}, left). 

The comparison of our three in-context learning strategies (Figure~\ref{fig:llm_size_strategy}, right) shows that zero-shot, few-shot, and frame-guided prompts achieve substantially equal F-scores. A more in-depth analysis reveals a fundamental trade-off in prompt design: the zero-shot approach excels at precision ($+9$ points), while few-shot and frame-guided prompting dramatically improve recall by $+15$ and $+14$ points over the zero-shot setup. The injection of knowledge from FrameNet does not seem to have any effect on CT classification, suggesting that semantic frame information adds limited discriminative power beyond what in-context examples already convey.

\begin{figure*}[t]
    \centering
    \includegraphics[width=0.9\textwidth]{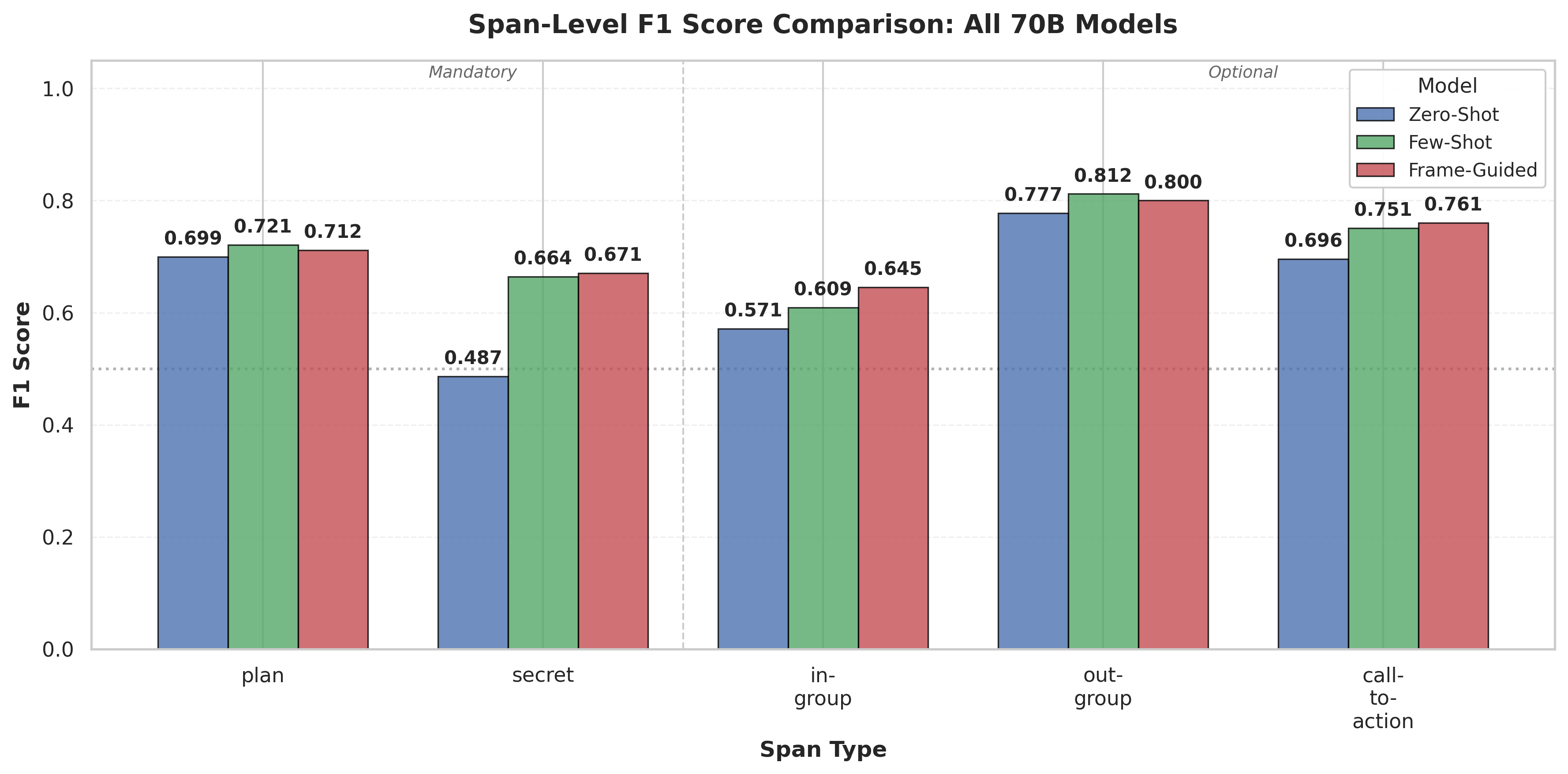}
    \caption{Span-Level F1 Score Comparison: All 70B Models}
    \label{fig:span_fscore_comparison}
\end{figure*}

The analysis of models performance in span classification shows more significant differences between in-context learning strategies. As it can be observed in Figure \ref{fig:span_fscore_comparison}, the zero-shot approach always achieves the worst performance in the detection of all the Conspiracy Frame elements. The highest differences can be observed in the detection of \textit{secret} ($-16$ and $-15$ F-score points) and \textit{call-to-action} ($-5$ and $-6$ points). Results also show a moderate divergence between the frame-guided and few-shot strategies, with the former achieving best results in the detection of \textit{in-group} ($+3.6$ points) and \textit{call-to-action} ($+1$ point); the latter obtaining a better F-score ($+1$) on the detection of \textit{out-group}.

\begin{figure*}[htbp]
    \centering
    \begin{subfigure}{0.45\textwidth}
        \centering
        \includegraphics[width=\linewidth]{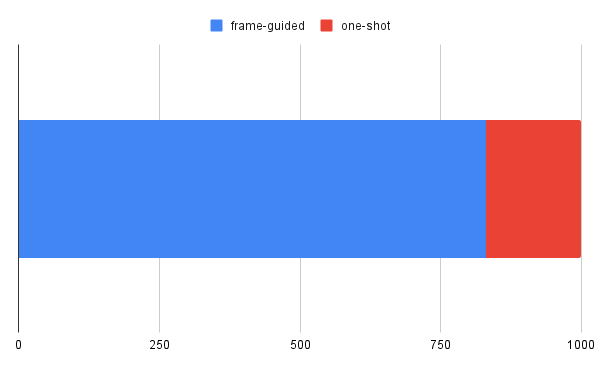}
        \caption{Classification Results}
        \label{fig:first}
    \end{subfigure}
    \hfill
    \begin{subfigure}{0.45\textwidth}
        \centering
        \includegraphics[width=\linewidth]{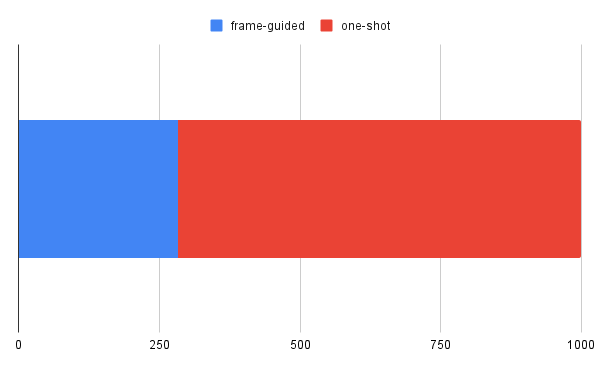}
        \caption{Span Results}
        \label{fig:second}
    \end{subfigure}
    \caption{The ELO scores of the Few-Shot and Frame-Guided settings. Figure on the left shows the comparison between of the two approaches in CT recognition; Figure on the right on span detection.}
    \label{fig:elo}
\end{figure*}

\paragraph{Models performance on out-of-domain data.} The second part of the experiment involved the evaluation of LLMs in the classification and span detection of CTs on $500$ unlabeled messages from different Telegram channels, as explained in Section \ref{sss:second}. LLMs were evaluated through an experimental setup based on the ELO system rating (Section \ref{sss:metrics}), which was repeated $1,000$ times to ensure robustness. In this setting, we tested the few-shot and the frame-guided approaches with Llama-3.3 70B and 8B. 

The comparison of models by size reveals systematic higher performances of bigger LLMs. The two settings based on Llama-3.3 70B obtains a highest ELO ranking in $998$ out of $1,000$ iterations in the classification task and $982$ out of $1,000$ in the span detection task.  

We performed the same evaluation between frame-guided and few-shot predictions of Llama-3.3 70B to evaluate the impact of frames in CT classification and span detection. As it is possible to observe in Figure \ref{fig:elo}, results appear to be mixed. The frame-guided approach outperforms the few-shot in text classification ($830/1,000$), but falls behind the few-shot in span detection ($213/1,000$).

We further investigated these results with an error analysis of CT spans based on the writing of explanations grounded on semi-structured templates. Two members of the research team reviewed the $60$ cases in which one detected span was preferred over another and wrote an explanation based on the template 'I believe that the frame-guided output was better/worst because [...]'. 

From this analysis, distinct patterns emerged governing the preference for one approach over the other: \textbf{span completeness} and \textbf{structural precision}.

Annotators consistently preferred the model that captured the full extent of the narrative, including specific mechanisms, targets, or consequences, over those that extracted fragmented keywords. This pattern emerged in both directions. The frame-guided model was preferred when it identified multiple relevant spans or detailed mechanisms: for instance,  in a text about 5G-activated nanopathogens, the frame-guided output captured ``can be activated by an 18 gigahertz signal on the 5G network'', while the baseline only extracted ``activating these dormant pathogens''. Similarly, the frame-guided model successfully identified both ``combating neo-Nazism'' and ``denazification'' as plan elements where the baseline captured only the former.
Conversely, the few-shot model was preferred when it captured a comprehensive causal statement that the frame-guided model truncated. In a text about COVID-19 origins, the baseline captured both ``secretly commissioned the paper'' and ``censoring discussions on the lab leak'', while the frame-guided model extracted only the first action; additionally, in the same text, the frame-guided model conflated plan and secret by using ``secretly commissioned'' for both, while the baseline correctly separated ``private messages'' as the secret and ``secretly commissioned the paper'' as the plan. In another case, the baseline included ``recruiting prostitutes'' alongside ``human trafficking'', which the frame-guided model omitted.

Second, a trade-off was observed between semantic depth and syntactic precision. The frame-guided model excelled in resolving ambiguity, correctly identifying implicit \textit{secret} that the few-shot approach reduced to meaningless adjectives. For instance, in a text about vaccine propaganda, the frame-guided model extracted ``trying to put in your head'' as a meaningful secret describing manipulation, while the baseline extracted only the word ``captured'', a fragment that loses semantic clarity in isolation despite referencing ``captured public health agencies''. However, this sensitivity sometimes led to false positives. In simpler texts, the few-shot model was frequently preferred for its precision, avoiding the inclusion of unnecessary auxiliary verbs or non-conspiratorial context that the frame-guided model forced into the template.

\subsection{Qualitative Analysis} \label{ss:qualitative}

The qualitative analysis of frames and their distribution considers human-annotated spans \textit{vs.} spans detected by Llama 3.3 70B in a few-shot setting to observe key similarities and differences with a model that was not explicitly prompted with semantic frames. 

A first observation emerging from the analysis shows that frames that appear to be coherent with the core elements of the Conspiracy Frame - \textsc{Execute\_Plan}\footnote{All the definitions of frames are accessible at the following url: \url{https://framenet.icsi.berkeley.edu/frames/}} (An Agent acts according to a Plan, carrying it out (or taking steps to carry it out) and \textsc{Secrecy\_Status} (A Phenomenon, which may be an activity, state or object, is purposefully hidden from the awareness of a potential Cognizer) are not among the $10$ most recurring frames either among human and LLM spans, which share frames that are more focused on conspiratorial narratives.

The frequent activation of the frames and \textsc{Political\_locales}, which situate conspiratorial action within institutional and geopolitical contexts (represented by lexical units such as `government', ‘state’, ‘facility’, and ‘work’), frame plans as exercises of organized power. The distinctly disruptive and anti-authoritarian drive of CTs is reflected here on the high recurrence of these frames.

Furthermore, CTs’ negative axiology \cite{[19]} is showcased by the fact that the frame \textsc{Statement} is the most recurring one in both human and LLM annotations. Here, terms such as `tell’, `report’, and `declare’ trigger conspiratorial plans, in a way that mainstream statements are thus seen as inherently misleading. This supports the view of CTs as counter-narratives formulated in opposition to official histories \cite{[20]}.


Two frames characterizing spans detected by LLMs are \textsc{Ingest\_substance} (take, use, shoot) and \textsc{Firefighting} (fight, control, attack). These frames point to how the rhetoric of CTs commonly work as metaphors for invasion, contagion, addiction, and war \cite{[18]}. The lexical realization of these frames contribute to the construction of conspiratorial plans as existential threats, amplifying their perceived severity.


Conversely, the recurrence of the frame \textsc{Calendric\_unit} in human-annotated spans points to the importance of the temporal dimension for CT rhetoric, where the discourse constructs a danger of the present, connected to scars from the past and linked to dreaded future scenarios \cite{[10]}. 

In sum, the most recurring frames emerging from human-annotated and LLM-detected spans point to how the operation of conspiracy theories does not depend so much on specific topic of interests, but on higher order beliefs that are partially aligned with FrameNet. This generalization is not applied with more abstract conceptualizations of \textit{secret} and \textit{plan}, showing the need of a better connection between CT detection and frame-semantics in order to capture more nuanced characteristics of conspiracy narratives.   

\section{Conclusion and Future Work} \label{sec:conclusion}
In this paper we presented the Conspiracy Frame: a fine-grained semantic representation of conspiratorial narratives derived from frame-semantics and semiotics, which inspired the creation of \textbf{Con}spiracy \textbf{Fra}mes (Con.Fra.) dataset: a corpus of Telegram messages annotated at span-level. This set of resources has been exploited to investigate the contribution of frame-semantics in enabling more generalizable and explainable understanding of Conspiracy Theories. Results show that while the augmentation of in-context learning approaches with frames does not give clear advantages, frames are important to explain the structure of conspiratorial narratives.
Future work will explore deeper integration of frame-semantics by fine-tuning or pretraining LLMs on knowledge graphs constructed from retrieved FrameNet frames. Such knowledge enhanced models could internalize frame semantics as structured relational knowledge, moving beyond surface-level frame hints in prompts toward deep integration of frame-based reasoning. 
By representing frames as nodes in a knowledge graph with typed relationships (e.g., frame-to-frame relations, frame-to-word mappings, frame-to-domain connections), we can enable models to learn and traverse richer semantic representations during inference. 

\section*{Limitations}
Since the focus of this work is on the impact of frame semantics on CT understanding, its experimental setting is limited to the Llama models. Choosing to focus on a single family of LLMs narrows the analysis of models performance. However, a systematic benchmarking of LLMs and SLMs on CT detection is out of the scope of the paper.

The annotation process showed cases of low agreement among participants. While this is a common issue traceable in other subjective phenomenons, we mitigate this aspect by releasing disaggregated annotations. This might also support the study of Human Label Variation in a very relevant field of research that is CT detection.

The lexical-based approach that we adopted to map Con.Fra with FrameNet was carefully conceived to reduce noise in the mapping but cannot resolve potential issues related to lexical ambiguity. Acknowledging this limitation, we performed manual checks of the resulting mapping. However, more sophisticated approaches to properly integrate semantic frames in CT detection are needed and will be the focus of our future work.

\section*{Ethical Considerations}
Studying CTs expose researchers and users to potential harms. For this reason, we followed a strict protocol to anonymize collected data, annotate them, and protect researchers from potential negative behaviors acted by activists belonging to the studied Telegram channels. The protocol was validated by the Ethical Committee of one of our researchers' institution.

The dataset annotation did not rely on crowdworkers. Students and researchers participated voluntarily to this initiative for their interest in the topic. We still shared with them terms and conditions of their participation, explaining their right to opt-out from the annotation anytime. We properly anonymized annotators identities, according to GDPR standards and warn them of the potential presence of offensive and discriminatory contents in data to annotate.

\bibliography{custom}

\appendix
\section{Annotation Guidelines} \label{app:guidelines}
These are the guidelines with which human annotators were presented in order to carry out the annotation task:

\textbf{Is the text conspiratorial?}

If YES: Carry on with annotation.

\textbf{If NO: Does the text contain a \textit{statement that may support} a conspiracy theory?}

If NO: Mark as ‘no’ and skip to the next text.

If YES: Mark as ‘yes’ and skip to the next text.

\begin{itemize}
\itemsep=-0.5em
\item The text IS conspiratorial if there is at least something indicating a plan/event that is the fault of an out-group (which may be only implied) and harms an in-group (may also just be implied). 
\item The plan part has to be identifiable for it to be a conspiracy theory text.
\item If there are only non-core elements (especially just a call-to-action), the text is NOT conspiratorial.
\item The text supports a conspiracy theory if it contains a statement that can be used as “proof” for the existence of a conspiratorial plan.
\end{itemize}

\textbf{Definition:}

An \colorbox{cbOrange}{event} is the result of a \colorbox{cbBlue}{group of people} acting in \colorbox{cbGreen}{secret} according to a \colorbox{cbOrange}{Plan}, causing harm to a \colorbox{cbPurple}{group of victims}. If not stopped, this plan will lead to \colorbox{cbOrange}{catastrophe}, therefore \colorbox{cbYellow}{actions need to be taken}.

\textbf{Hypothetical examples:}

\vspace{0.5em}

\noindent \textbf{Example 1:} \\
\colorbox{cbOrange}{Forced vaccination is being institutionalized} by the \colorbox{cbBlue}{authorities}, to \colorbox{cbOrange}{profit from and control} the \colorbox{cbPurple}{population} \colorbox{cbGreen}{under the guise} of public health. We need to \colorbox{cbYellow}{stand against this oppression}.

\vspace{0.8em}

\noindent \textbf{Example 2:} \\
\colorbox{cbOrange}{Multicultural policies increase immigration}, serving to \colorbox{cbOrange}{replace} all \colorbox{cbPurple}{Christian white people} with \colorbox{cbBlue}{Muslims}, and \colorbox{cbOrange}{turning Europe into a colony of Islam}. You need to \colorbox{cbYellow}{stand up and join the fight} against invasion!

\vspace{0.8em}

\noindent \textbf{Example 3:} \\
\colorbox{cbPurple}{White folk} is living under the thumb of a \colorbox{cbBlue}{government} that is \colorbox{cbOrange}{catering to minorities} and \colorbox{cbOrange}{hostile} to \colorbox{cbPurple}{us}. Our hard-earned \colorbox{cbOrange}{money is being shipped overseas to fund} \colorbox{cbOrange}{money laundering schemes} \colorbox{cbGreen}{packaged as humanitarian aid}. Our borders are wide open \colorbox{cbOrange}{importing} \colorbox{cbBlue}{people who have demonstrated hostility} to \colorbox{cbPurple}{us, and our families}. We are heading for \colorbox{cbOrange}{certain demise}.

\vspace{1.5em}

\textbf{Core elements:}

\noindent \colorbox{cbOrange}{\textbf{Plan / event}} \\
The circumstances and/or action(s) being taken to achieve an evil end. The ultimate objective of the \colorbox{cbBlue}{out-group} that harms the \colorbox{cbPurple}{in-group} and leads to catastrophe. It is conspiratorial.

\vspace{0.5em}

\noindent \textbf{Example 1:} \\
\colorbox{cbBlue}{Public schools} \colorbox{cbOrange}{indoctrinate} \colorbox{cbPurple}{our children} \colorbox{cbOrange}{with marxist doctrine}.

\vspace{0.5em}

\noindent \textbf{Example 2:} \\
Yesterday, \colorbox{cbPurple}{a white man} was \colorbox{cbOrange}{assaulted} by a \colorbox{cbBlue}{group of non-white invaders}.

\vspace{1em}

\noindent \colorbox{cbGreen}{\textbf{Secret}} \\
Gives the connotation of secrecy to the \colorbox{cbOrange}{plan}.

\vspace{0.5em}

\noindent \textbf{Example:} \\
\colorbox{cbGreen}{The real reason} for "the invasion" of the USSR was an attempt to \colorbox{cbYellow}{prevent} all \colorbox{cbPurple}{Europe} from \colorbox{cbOrange}{becoming overrun by the red army}.

\vspace{1.5em}

\textbf{Non-core elements:}

\noindent \colorbox{cbBlue}{Out-group}: The enemy - “them” \\
\colorbox{cbPurple}{In-group}: The victims - “us”

\vspace{0.5em}

\noindent \textbf{Example:} \\
The \colorbox{cbBlue}{International Jewry} keep \colorbox{cbOrange}{starting wars, but they never fight in it}, only the \colorbox{cbPurple}{White Man} \colorbox{cbOrange}{die in the process}.

\vspace{1em}

\noindent \colorbox{cbYellow}{\textbf{Call-to-action}} \\
What should be done to avoid the \colorbox{cbOrange}{catastrophe}.

\vspace{0.5em}

\noindent \textbf{Example:} \\
\colorbox{cbPurple}{Our race} is under \colorbox{cbOrange}{threat of absolute extinction}. Either we \colorbox{cbYellow}{save it} or \colorbox{cbOrange}{die}. \colorbox{cbYellow}{Share this message} around to wake your neighbours.

If conspiratorial = YES, then the mandatory elements are either [plan/event] AND/OR [secret], one must be identified, others are optional.
\onecolumn
\section{Prompts} \label{app:prompt}

\subsection{Zero-Shot}
\small

\textbf{Role.} You are an expert annotator. For each Telegram message: return only JSON.

\vspace{0.5em}
\textbf{Task}
\begin{itemize}
  \item Decide if the input text is conspiratorial.
  \item Provide a short rationale summarizing your decision.
  \item Give a confidence score between 0 and 1 (inclusive, use decimals).
  \item If conspiratorial, extract labeled spans.
\end{itemize}

\vspace{0.5em}
\textbf{Definition}
\begin{itemize}
  \item \textbf{Conspiracy theory}: An event is the result of a group acting in secret according to a plan, causing harm to a group of victims. If not stopped, it leads to catastrophe, so action is urged.
  \item A text is conspiratorial if there is at least one identifiable plan/event attributed to an out-group (may be implied) that harms an in-group (may be implied).
  \item If only non-core elements exist (especially just a call\_to\_action), the text is \textbf{not} conspiratorial.
\end{itemize}

\vspace{0.5em}
\textbf{Core Requirement}
\begin{itemize}
  \item When \texttt{is\_conspiratorial} is true, include at least one span labeled \texttt{plan\_event} or \texttt{secret}.
\end{itemize}

\vspace{0.5em}
\textbf{Span Labels}
\begin{itemize}
  \item \texttt{plan\_event} (core): action/event/circumstance that constitutes the harmful plan.
  \item \texttt{secret} (core): secrecy/hidden coordination cues (e.g., secret, covert, hidden agenda).
  \item \texttt{out\_group} (optional): alleged perpetrators/enemy.
  \item \texttt{in\_group} (optional): victims/``us''.
  \item \texttt{call\_to\_action} (optional): imperative/prescription to counter the plan.
\end{itemize}

\vspace{0.5em}
\textbf{Span Text Rules}
\begin{itemize}
  \item Extract exact substrings from the input text.
  \item Do NOT include surrounding quotes or trailing punctuation in span text.
  \item Span text must be continuous.
  \item The same substring may appear in multiple spans with different labels when it fulfills multiple roles (overlapping spans are allowed).
\end{itemize}

\vspace{0.5em}
\textbf{Output JSON Schema}
\begin{verbatim}
{
  "is_conspiratorial": boolean,
  "rationale_short": string,
  "confidence": number,
  "spans": [
    {
      "label": "plan_event" | "secret" | "out_group" | "in_group" | "call_to_action",
      "text": string
    }
  ]
}
\end{verbatim}

\vspace{0.5em}
\textbf{Input}
\begin{verbatim}
{{INPUT_TEXT}}
\end{verbatim}

\vspace{0.5em}
\textbf{Output}
Return only the JSON.

\subsection{Few-shot}
\small

\textbf{Role.} You are an expert annotator. For each Telegram message: return only JSON.

\vspace{0.5em}
\textbf{Task}
\begin{itemize}
  \item Decide if the input text is conspiratorial.
  \item Provide a short rationale summarizing your decision.
  \item Give a confidence score between 0 and 1 (inclusive, use decimals).
  \item If conspiratorial, extract labeled spans.
\end{itemize}

\vspace{0.5em}
\textbf{Definition}
\begin{itemize}
  \item \textbf{Conspiracy theory}: An event is the result of a group acting in secret according to a plan, causing harm to a group of victims. If not stopped, it leads to catastrophe, so action is urged.
  \item A text is conspiratorial if there is at least one identifiable plan/event attributed to an out-group (may be implied) that harms an in-group (may be implied).
  \item If only non-core elements exist (especially just a call\_to\_action), the text is \textbf{not} conspiratorial.
\end{itemize}

\vspace{0.5em}
\textbf{Core Requirement}
\begin{itemize}
  \item When \texttt{is\_conspiratorial} is true, include at least one span labeled \texttt{plan\_event} or \texttt{secret}.
\end{itemize}

\vspace{0.5em}
\textbf{Span Labels}
\begin{itemize}
  \item \texttt{plan\_event} (core): action/event/circumstance that constitutes the harmful plan.
  \item \texttt{secret} (core): secrecy/hidden coordination cues (e.g., secret, covert, hidden agenda).
  \item \texttt{out\_group} (optional): alleged perpetrators/enemy.
  \item \texttt{in\_group} (optional): victims/``us''.
  \item \texttt{call\_to\_action} (optional): imperative/prescription to counter the plan.
\end{itemize}

\vspace{0.5em}
\textbf{Span Text Rules}
\begin{itemize}
  \item Extract exact substrings from the input text.
  \item Do NOT include surrounding quotes or trailing punctuation in span text.
  \item Span text must be continuous.
  \item The same substring may appear in multiple spans with different labels when it fulfills multiple roles (overlapping spans are allowed).
\end{itemize}

\vspace{0.5em}
\textbf{Output JSON Schema}
\begin{verbatim}
{
  "is_conspiratorial": boolean,
  "rationale_short": string,
  "confidence": number,
  "spans": [
    {
      "label": "plan_event" | "secret" | "out_group" | "in_group" | "call_to_action",
      "text": string
    }
  ]
}
\end{verbatim}

\vspace{0.5em}
\textbf{Few-shot Examples}

\textbf{Example 1 (Conspiratorial — plan\_event + out\_group + in\_group)}
\begin{verbatim}
Input: 
Influence Operation Relied on Influencers, AI-Generated Content, Paid Social Media Advertisements, and Social Media Accounts to Drive Internet Traffic to Cybersquatted and Other Domains

The Justice Department today announced the ongoing seizure of 32 internet domains used in Russian government-directed foreign malign influence campaigns colloquially referred to as "Doppelganger," in violation of U.S. money laundering and criminal trademark laws. As alleged in an unsealed affidavit, the Russian companies Social Design Agency (SDA), Structura National Technology (Structura), and ANO Dialog, operating under the direction and control of the Russian Presidential Administration, and in particular First Deputy Chief of Staff of the Presidential Executive Office Sergei Vladilenovich Kiriyenko, used these domains, among others, to covertly spread Russian government propaganda with the aim of reducing international support for Ukraine, bolstering pro-Russian policies and interests, and influencing voters in U.S. and foreign elections, including the U.S. 2024 Presidential Election.

Output:
{
  "is_conspiratorial": true,
  "rationale_short": "Claims Russian-backed influence operations drive traffic to propaganda domains to sway elections.",
  "confidence": 0.83,
  "spans": [
    { "label": "plan_event", "text": "Drive Internet Traffic to Cybersquatted and Other Domains" },
    { "label": "out_group", "text": "Russian government" },
    { "label": "in_group", "text": "U.S." }
  ]
}
\end{verbatim}

\textbf{Example 2 (Conspiratorial — multiple plan\_events + secret + out\_group)}
\begin{verbatim}
Input:
Jewish German Government -

"You can be thankful we let the third world in making your cities more unsafe... you can be thankful we turn your children gay and punish you for speaking out against immigration and the holohoax.  Now we will light up the Brandenburg Gate with the satanic star of Remphan to show you who rules over you.  Please be stupid and buy into our tears for help one more time."

People everywhere will start learning what these savages do to the Palestinians. Support for Israel will only decline until they seem to be the last ones cheering themselves on.

Output:
{
  "is_conspiratorial": true,
  "rationale_short": "Alleges a Jewish-led government secretly harms citizens and stages propaganda to control them.",
  "confidence": 0.80,
  "spans": [
    { "label": "plan_event", "text": "punish you for speaking out against immigration and the holohoax" },
    { "label": "plan_event", "text": "light up the Brandenburg Gate with the satanic star of Remphan" },
    { "label": "secret", "text": "buy into" },
    { "label": "out_group", "text": "Jewish German Government" },
    { "label": "out_group", "text": "savages" }
  ]
}
\end{verbatim}

\textbf{Example 3 (Conspiratorial — plan\_event + secret, policy focus)}
\begin{verbatim}
Input:
  "Courts will sit for 24 hours to fast-track sentencing under government plans to crack down on far-Right riots that swept Britain on Saturday"

 : Funny they can't do the same with illegals.

Output:
{
  "is_conspiratorial": true,
  "rationale_short": "Claims the government secretly targets far-right activists while protecting illegal immigrants.",
  "confidence": 0.72,
  "spans": [
    { "label": "plan_event", "text": "government plans to crack down on far-Right riots" },
    { "label": "plan_event", "text": "can't do the same with illegals" },
    { "label": "secret", "text": "they can't do the same with illegals" },
    { "label": "in_group", "text": "far-Right riots" }
  ]
}
\end{verbatim}

\textbf{Example 4 (Conspiratorial — plan\_event + in\_group + call\_to\_action)}
\begin{verbatim}
Input:
Texans are known for their resilience and ability to persevere in the face of adversity. From enduring harsh weather conditions like hurricanes and droughts to facing racial demographic challenges, Texans have shown time and time again that they are a tough and resilient people. Their strong sense of community and willingness to help one another in times of need is a testament to their unwavering spirit. Despite facing numerous obstacles, Texans always find a way to come together, support each other, and rebuild stronger than before, as evidenced by the reformation of WLM Texas. This resilience is ingrained in the Texan spirit and serves as a source of inspiration for others facing similar challenges. 
If you're a White man or woman in the state of Texas, reach out. The community and system of support you've been looking for is right here.

Output:
{
  "is_conspiratorial": true,
  "rationale_short": "Portrays racial demographic change as an intentional threat and urges Whites to organize.",
  "confidence": 0.69,
  "spans": [
    { "label": "plan_event", "text": "racial demographic challenges" },
    { "label": "in_group", "text": "White man or woman" },
    { "label": "call_to_action", "text": "always find a way to come together, support each other, and rebuild stronger than before" }
  ]
}
\end{verbatim}

\textbf{Example 5 (Not conspiratorial — neutral coordination request)}
\begin{verbatim}
Input:
Hey guys, we're trying to get this community noted. 

Can everyone please tag   in the comments and correct us.

Output:
{
  "is_conspiratorial": false,
  "rationale_short": "Asks followers to tag accounts; no secret plan or coordinated harm is described.",
  "confidence": 0.24,
  "spans": []
}
\end{verbatim}

\textbf{Example 6 (Not conspiratorial — supportive tone only)}
\begin{verbatim}
Input:
Someone should comment this under Elon Musks Comment he made about Haiti.

In fact... most of you should clip parts of this and paste this link below so everyone can watch it all over twitter. Everyone is wondering why they eat people... maybe this will enlighten them a bit. Scientific brain IQ and brain measurements are also useful to post.

Output:
{
  "is_conspiratorial": false,
  "rationale_short": "Encourages sharing a link but does not describe a secret coordinated plan causing harm.",
  "confidence": 0.31,
  "spans": []
}
\end{verbatim}

\vspace{0.5em}
\textbf{Input}
\begin{verbatim}
{{INPUT_TEXT}}
\end{verbatim}

\vspace{0.5em}
\textbf{Output}
Return only the JSON.

\subsection{Frame-guided}
\small

\textbf{Role.} You are an expert annotator. For each Telegram message: return only JSON.

\vspace{0.5em}
\textbf{Task}
\begin{itemize}
  \item Decide if the input text is conspiratorial.
  \item Provide a short rationale summarizing your decision.
  \item Give a confidence score between 0 and 1 (inclusive, use decimals).
  \item If conspiratorial, extract labeled spans.
\end{itemize}

\vspace{0.5em}
\textbf{Definition}
\begin{itemize}
  \item \textbf{Conspiracy theory}: An event is the result of a group acting in secret according to a plan, causing harm to a group of victims. If not stopped, it leads to catastrophe, so action is urged.
  \item A text is conspiratorial if there is at least one identifiable plan/event attributed to an out-group (may be implied) that harms an in-group (may be implied).
  \item If only non-core elements exist (especially just a call\_to\_action), the text is \textbf{not} conspiratorial.
\end{itemize}

\vspace{0.5em}
\textbf{Core Requirement}
\begin{itemize}
  \item When \texttt{is\_conspiratorial} is true, include at least one span labeled \texttt{plan\_event} or \texttt{secret}.
\end{itemize}

\vspace{0.5em}
\textbf{Span Labels}
\begin{itemize}
  \item \texttt{plan\_event} (core): action/event/circumstance that constitutes the harmful plan.
  \item \texttt{secret} (core): secrecy/hidden coordination cues (e.g., secret, covert, hidden agenda).
  \item \texttt{out\_group} (optional): alleged perpetrators/enemy.
  \item \texttt{in\_group} (optional): victims/``us''.
  \item \texttt{call\_to\_action} (optional): imperative/prescription to counter the plan.
\end{itemize}

\vspace{0.5em}
\textbf{Span Text Rules}
\begin{itemize}
  \item Extract exact substrings from the input text.
  \item Do NOT include surrounding quotes or trailing punctuation in span text.
  \item Span text must be continuous.
  \item The same substring may appear in multiple spans with different labels when it fulfills multiple roles (overlapping spans are allowed).
\end{itemize}

\vspace{0.5em}
\textbf{Frame Hints}
Some examples include frame hints for core spans (\texttt{plan\_event}, \texttt{secret}). These are provided to give you extra context but should \textbf{not} be added to the JSON output.

\vspace{0.5em}
\textbf{Output JSON Schema}
\begin{verbatim}
{
  "is_conspiratorial": boolean,
  "rationale_short": string,
  "confidence": number,
  "spans": [
    {
      "label": "plan_event" | "secret" | "out_group" | "in_group" | "call_to_action",
      "text": string
    }
  ]
}
\end{verbatim}

\vspace{0.5em}
\textbf{Few-shot Examples}

\textbf{Example 1 (Conspiratorial — plan\_event + out\_group + in\_group)}
\begin{verbatim}
Input: 
Influence Operation Relied on Influencers, AI-Generated Content, Paid Social Media Advertisements, and Social Media Accounts to Drive Internet Traffic to Cybersquatted and Other Domains

The Justice Department today announced the ongoing seizure of 32 internet domains used in Russian government-directed foreign malign influence campaigns colloquially referred to as "Doppelganger," in violation of U.S. money laundering and criminal trademark laws. As alleged in an unsealed affidavit, the Russian companies Social Design Agency (SDA), Structura National Technology (Structura), and ANO Dialog, operating under the direction and control of the Russian Presidential Administration, and in particular First Deputy Chief of Staff of the Presidential Executive Office Sergei Vladilenovich Kiriyenko, used these domains, among others, to covertly spread Russian government propaganda with the aim of reducing international support for Ukraine, bolstering pro-Russian policies and interests, and influencing voters in U.S. and foreign elections, including the U.S. 2024 Presidential Election.

Frame hints:
- plan_event: Self_motion, Subjective_influence, Bringing, Cause_motion, Operate_vehicle

Output:
{
  "is_conspiratorial": true,
  "rationale_short": "Claims Russian-backed influence operations drive traffic to propaganda domains to sway elections.",
  "confidence": 0.83,
  "spans": [
    { "label": "plan_event", "text": "Drive Internet Traffic to Cybersquatted and Other Domains" },
    { "label": "out_group", "text": "Russian government" },
    { "label": "in_group", "text": "U.S." }
  ]
}
\end{verbatim}

\textbf{Example 2 (Conspiratorial — multiple plan\_events + secret + out\_group)}
\begin{verbatim}
Input:
Jewish German Government -

"You can be thankful we let the third world in making your cities more unsafe... you can be thankful we turn your children gay and punish you for speaking out against immigration and the holohoax.  Now we will light up the Brandenburg Gate with the satanic star of Remphan to show you who rules over you.  Please be stupid and buy into our tears for help one more time."

People everywhere will start learning what these savages do to the Palestinians. Support for Israel will only decline until they seem to be the last ones cheering themselves on.

Frame hints:
- plan_event: Rewards_and_punishments, Statement, Chatting, Color_qualities, Setting_fire
- secret: Commerce_buy

Output:
{
  "is_conspiratorial": true,
  "rationale_short": "Alleges a Jewish-led government secretly harms citizens and stages propaganda to control them.",
  "confidence": 0.80,
  "spans": [
    { "label": "plan_event", "text": "punish you for speaking out against immigration and the holohoax" },
    { "label": "plan_event", "text": "light up the Brandenburg Gate with the satanic star of Remphan" },
    { "label": "secret", "text": "buy into" },
    { "label": "out_group", "text": "Jewish German Government" },
    { "label": "out_group", "text": "savages" }
  ]
}
\end{verbatim}

\textbf{Example 3 (Conspiratorial — plan\_event + secret, policy focus)}
\begin{verbatim}
Input:
  "Courts will sit for 24 hours to fast-track sentencing under government plans to crack down on far-Right riots that swept Britain on Saturday"

 : Funny they can't do the same with illegals.

Frame hints:
- plan_event: Leadership, Organization, Project, Purpose, Making_arrangements
- secret: Intentionally_act, Legality, Identicality, Thriving, Intentionally_affect

Output:
{
  "is_conspiratorial": true,
  "rationale_short": "Claims the government secretly targets far-right activists while protecting illegal immigrants.",
  "confidence": 0.72,
  "spans": [
    { "label": "plan_event", "text": "government plans to crack down on far-Right riots" },
    { "label": "plan_event", "text": "can't do the same with illegals" },
    { "label": "secret", "text": "they can't do the same with illegals" },
    { "label": "in_group", "text": "far-Right riots" }
  ]
}
\end{verbatim}

\textbf{Example 4 (Conspiratorial — plan\_event + in\_group + call\_to\_action)}
\begin{verbatim}
Input:
Texans are known for their resilience and ability to persevere in the face of adversity. From enduring harsh weather conditions like hurricanes and droughts to facing racial demographic challenges, Texans have shown time and time again that they are a tough and resilient people. Their strong sense of community and willingness to help one another in times of need is a testament to their unwavering spirit. Despite facing numerous obstacles, Texans always find a way to come together, support each other, and rebuild stronger than before, as evidenced by the reformation of WLM Texas. This resilience is ingrained in the Texan spirit and serves as a source of inspiration for others facing similar challenges. 
If you're a White man or woman in the state of Texas, reach out. The community and system of support you've been looking for is right here.

Frame hints:
- plan_event: Statement, Competition, Difficulty

Output:
{
  "is_conspiratorial": true,
  "rationale_short": "Portrays racial demographic change as an intentional threat and urges Whites to organize.",
  "confidence": 0.69,
  "spans": [
    { "label": "plan_event", "text": "racial demographic challenges" },
    { "label": "in_group", "text": "White man or woman" },
    { "label": "call_to_action", "text": "always find a way to come together, support each other, and rebuild stronger than before" }
  ]
}
\end{verbatim}

\textbf{Example 5 (Not conspiratorial — neutral coordination request)}
\begin{verbatim}
Input:
Hey guys, we're trying to get this community noted. 

Can everyone please tag   in the comments and correct us.

Output:
{
  "is_conspiratorial": false,
  "rationale_short": "Asks followers to tag accounts; no secret plan or coordinated harm is described.",
  "confidence": 0.24,
  "spans": []
}
\end{verbatim}

\textbf{Example 6 (Not conspiratorial — supportive tone only)}
\begin{verbatim}
Input:
Someone should comment this under Elon Musks Comment he made about Haiti.

In fact... most of you should clip parts of this and paste this link below so everyone can watch it all over twitter. Everyone is wondering why they eat people... maybe this will enlighten them a bit. Scientific brain IQ and brain measurements are also useful to post.

Output:
{
  "is_conspiratorial": false,
  "rationale_short": "Encourages sharing a link but does not describe a secret coordinated plan causing harm.",
  "confidence": 0.31,
  "spans": []
}
\end{verbatim}

\vspace{0.5em}
\textbf{Input}
\begin{verbatim}
{{INPUT_TEXT}}
\end{verbatim}

\vspace{0.5em}
\textbf{Output}
Return only the JSON.

\twocolumn

\end{document}